\documentclass[preprint,review,12pt]{elsarticle}

\usepackage{amssymb}
\usepackage{amsmath,amsfonts}

\usepackage{lineno}
\modulolinenumbers[5]
\usepackage{hyperref}
\usepackage{graphics}
\usepackage{float}
\usepackage{booktabs}
\usepackage{adjustbox}
\usepackage{multirow}

\journal{Information Fusion}

\pdfoutput=1

\begin{document}

\begin{frontmatter}



\title{Towards Deeper and Better Multi-view Feature Fusion for 3D Semantic Segmentation}


\author[1]{Chaolong Yang} 
\ead{chaolong.yang@dukekunshan.edu.cn}

\author[2]{Yuyao Yan}
\ead{joshuayyy@gmail.com}

\author[1]{Weiguang Zhao}
\ead{weiguang.zhao@dukekunshan.edu.cn}

\author[2]{Jianan Ye}
\ead{Jianan.Ye20@student.xjtlu.edu.cn}

\author[2]{Xi Yang}
\ead{xi.yang01@xjtlu.edu.cn}

\author[3]{Amir Hussain}
\ead{A.Hussain@napier.ac.uk}

\author[1]{Kaizhu Huang\corref{cor1}} 
\ead{kaizhu.huang@dukekunshan.edu.cn}

\address[1]{Data Science Research Center, Duke Kunshan University, Kunshan, 215316, China.}

\address[2]{Department of Electrical and Electronic Engineering, Xi'an Jiaotong-Liverpool University, Suzhou, 215123, China.}

\address[3]{School of Computing, Edinburgh Napier University, Edinburgh, EH11 4BN, UK.}


            

\cortext[cor1]{Corresponding author}

\begin{abstract}
3D point clouds are rich in geometric structure information, while 2D images contain important and continuous texture information. Combining 2D information to achieve better 3D semantic segmentation has become mainstream in 3D scene understanding. Albeit the success, it still remains elusive how to fuse and process the cross-dimensional features from these two distinct spaces. Existing state-of-the-art usually exploit bidirectional projection methods to align the cross-dimensional features and realize both 2D \& 3D semantic segmentation tasks. However, to enable bidirectional mapping, this framework  often requires a symmetrical 2D-3D network structure, thus limiting the network's flexibility. Meanwhile, such dual-task settings may distract the network easily and lead to over-fitting in the 3D segmentation task. 
As limited by the network's inflexibility, fused features can only pass through a decoder network, which affects model performance due to insufficient depth. To alleviate these drawbacks, in this paper, we argue that despite its simplicity, projecting unidirectionally multi-view 2D deep semantic features into the 3D space aligned with 3D deep semantic features could lead to better feature fusion. On the one hand, the unidirectional projection enforces our model focused more on the core task, i.e., 3D segmentation; on the other hand, unlocking the bidirectional  to unidirectional projection enables a deeper cross-domain semantic alignment and enjoys the flexibility to fuse better and complicated features from very different spaces. In joint 2D-3D approaches, our proposed method achieves superior performance on the ScanNetv2 benchmark for 3D semantic segmentation.

\end{abstract}


\begin{highlights}
\item We demonstrate that a unidirectional projection feature fusion scheme enables focus more on 3D semantic segmentation.
\item We design a novel framework based on semantic space alignment for 2D \& 3D information fusion.
\item Our framework reaches the state-of-the-art on ScanNetv2 and NYUv2 datasets.

\end{highlights}

\begin{keyword}
Point cloud, Semantic segmentation, Multi-view feature learning, Information fusion



\end{keyword}

\end{frontmatter}


\section{Introduction}
\label{sec:intro}
Semantic understanding of scenes is a core technology necessary in many fields, such as robot navigation, robotic arm grasping system, automatic driving system~\cite{FERNANDES2021161}, and medical diagnosis. Early researchers mainly analyze 2D images to achieve  semantic  scene understanding. Various 2D image scene understanding methodologies have boomed, including 2D image classification~\cite{szegedy2015going, he2016deep}, and semantic segmentation~\cite{yu2015multi, zhao2017pyramid, yao2021scaffold}. However, fixed-view 2D images suffer from object occlusion and missing spatial structure information. They  cannot provide better scene understanding for spatially location-sensitive downstream tasks. In contrast, 3D data, usually represented as a point cloud with unordered and irregular properties, offer a complete spatial structure without object occlusion. 

Traditional 2D neural network-based methodologies cannot be used directly to deal with 3D data. To this end, point-based~\cite{qi2017pointnet, qi2017pointnet++, wu2019pointconv, jiang2023pointgs} and voxel-based~\cite{graham20183d, choy20194d, zhao2022divide} neural networks have been explored for 3D point cloud recognition and understanding. Nevertheless, 3D point clouds are sparsely distributed in space, resulting in low resolution and a lack of rich texture information. Therefore, combining 2D images with detailed texture information and 3D point clouds with rich knowledge of geometric structures has been proposed as one promising solution~\cite{dai20183dmv, chiang2019unified, jaritz2019multi, kundu2020virtual, hu2021bidirectional} to understand complex scenes jointly. Extensive investigations have demonstrated that 2D \& 3D internal information are complementary, and 2D details can substantially improve the segmentation accuracy of 3D scenes.

\begin{figure}
	\centering
	\includegraphics[width=1\textwidth]{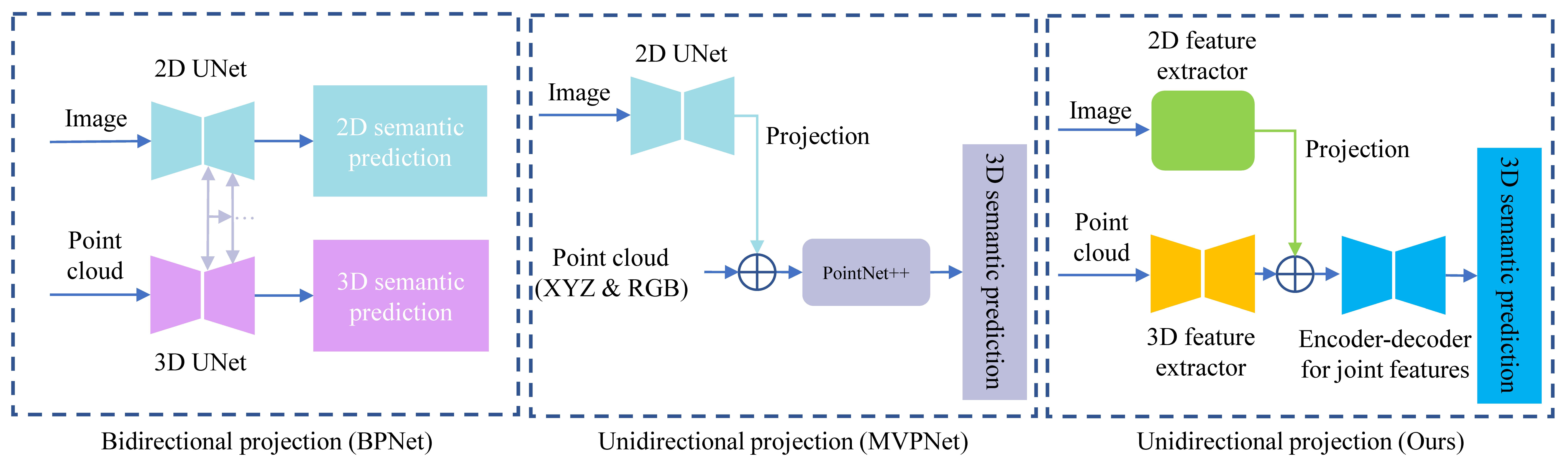}
	\caption{Comparison of bidirectional \& unidirectional projection}
	\label{fig:Framework Comparison of Bidirectional and Unidirectional Projection}
\end{figure}

\begin{figure}
	\centering
	\includegraphics[width=0.5\textwidth]{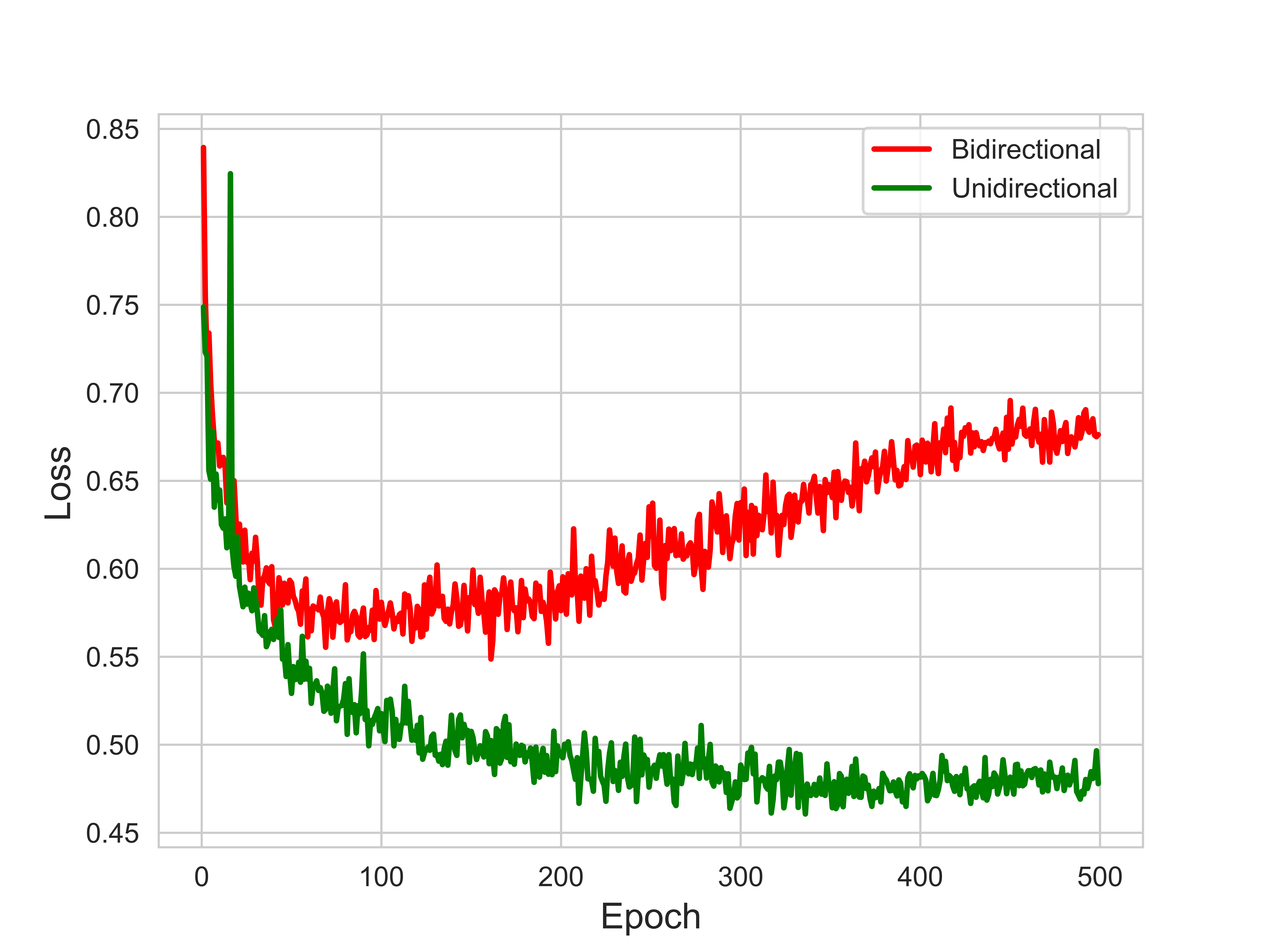}
	\caption{Validation loss for unidirectional \& bidirectional projection on validation set of the benchmark ScanNetv2 data}
	\label{fig:Validation Loss for Unidirectional & Bidirectional Projection}
\end{figure}

2D-3D fusion schemes for 3D semantic segmentation tasks can be divided into bidirectional and unidirectional projection.  Typically, network framework comparison between bidirectional projection and unidirectional projection can be seen in Figure~\ref{fig:Framework Comparison of Bidirectional and Unidirectional Projection}. The pioneering work of the bidirectional projection scheme, BPNet~\cite{hu2021bidirectional}, can enable 2D and 3D features to flow between networks. Nevertheless, in order to mutually fuse information from 2D to 3D and 3D to 2D, it usually has to exploit a symmetrical decoder network. This makes its framework less flexible and could not take advantage of network depth, thus limiting its performance. On the other hand, with the bidirectional information flow, it is often observed that the 2D semantic component, e.g. in the segmentation task may distract the network from the core 3D task, especially in complicated scenes. To illustrate, we implement the idea of bidirectional projection on our proposed unidirectional projection framework. Namely, 3D features are also projected into 2D space combined with 2D features and then input to a complete 2D encoder-decoder network. On a large-scale complex indoor scene ScanNetv2~\cite{dai2017scannet},  we compare in Figure~\ref{fig:Validation Loss for Unidirectional & Bidirectional Projection} the 3D semantic loss on the validation set for unidirectional and bidirectional projection ideas during model training.  Clearly, on the complicated scene of ScanNetv2, the bidirectional projection scheme causes distraction in the 3D task, where the loss goes up as the training continues. In comparison, the uni-projection implementation would lead to more stable performance with the focus mainly on the 3D task.



Motivated by these findings, in this paper, we argue that despite its simplicity projecting unidirectionally multi-view 2D deep semantic features into the 3D space aligned with 3D deep semantic features could lead to better feature fusion. Concretely, in the unidirectional projection solution, without the constraints of the bidirectional information flow, the network framework can be designed in a  way such that one can strengthen the backbone network deeper and more flexibly. As a result, it would be more convenient for the 2D-3D fused features to pass through a complete network of encoder-decoder, which is deep enough to allow the fused features to be fully learned. As such, the resulting network would offer more potential in fusing better 2D \& 3D information for downstream tasks like scene understanding.



There have been some previous unidirectional projection methods~\cite{chiang2019unified, jaritz2019multi} in the literature. These methods basically fuse 2D deep semantics and 3D shallow information (XYZ \& RGB) simply and then feed them into 3D backbone networks (see the middle graph in Figure~\ref{fig:Framework Comparison of Bidirectional and Unidirectional Projection}). Direct connection of deep and shallow semantics from different data domains could however result in the misalignment of the semantic space. To this end, we design a novel unidirectional projection framework, Deep Multi-view Fusion Network (DMF-Net) to more effectively fuse 2D \& 3D features (see the 
right graph in Figure~\ref{fig:Framework Comparison of Bidirectional and Unidirectional Projection}). The network consists of three main sub-networks. Namely, the 2D feature extractor extracts pixel-level texture details from multiple views, while the 3D feature extractor extracts voxel-level geometric structure information from the original point cloud. Subsequently, the multi-view 2D texture information can be back-projected to the 3D space and aligned with the original point cloud through the unidirectional projection module. Moreover, the adjacent 2D semantic features of each point in the original point cloud are found and summed up. It should be noted that we load the pre-trained 2D feature extraction module e.g. Swin-UNet~\cite{cao2021swin}, and freeze their parameters to make the model more focused on learning the 3D segmentation task. Finally, the fused 2D-3D features are fed into a 3D sparse convolution network to predict the semantic results of the 3D scene.

On the implementation front,  we evaluate our model on the task of 3D semantic segmentation on the indoor challenging dataset ScanNetv2~\cite{dai2017scannet}. DMF-Net achieves top performance for joint 2D-3D methods on the ScanNetv2 benchmark. Qualitative results also demonstrate the effectiveness of the fused 2D \& 3D information for 3D semantic segmentation tasks. In addition, DMF-Net is better in discriminating objects with little difference in shape thanks to 2D texture information, such as windows, walls, and pictures. To test the generalization ability of our model, we evaluate on another RGB-D dataset, NYUv2~\cite{silberman2012indoor}, and the results show that DMF-Net also achieves state-of-the-art performance on different datasets. 
Our contributions can be summarized as follows.
\begin{itemize}
    \item We argue that the unidirectional projection mechanism is not only more focused on 3D semantic understanding tasks than bidirectional projection but also facilitates deeper feature fusion. To this end, we design a method for uni-directional cross-domain semantic feature fusion to extract 2D \& 3D deep features for alignment simultaneously.

    \item We propose a novel framework named Deep Multi-view Fusion Network (DMF-Net) for 3D scene semantic understanding. For the joint 2D-3D approaches, DMF-Net obtains top mIOU performance on the 3D Semantic Label Benchmark of ScanNetv2~\cite{dai2017scannet}, while it reaches the state-of-the-art on NYUv2~\cite{silberman2012indoor} datasets.

    \item We demonstrate the flexibility of DMF-Net, where all backbone modules in the network framework can be replaced. Specifically, 3D semantic segmentation performance will be stronger with a powerful backbone. Compared with common U-Net34~\cite{ronneberger2015u}, the advanced Swin-Unet~\cite{cao2021swin} shows a relative 4.4\% improvement on mIOU in our framework.
    
    
	
	

\end{itemize}

\section{Related Work}
\label{sec:rel-work}

\subsection{2D Semantic Segmentation}
Image semantic segmentation has been significantly improved by the development of deep-learning~\cite{huang2019deep} models. In the field of 2D semantic segmentation, Fully Convolution Network (FCN)~\cite{long2015fully} is a landmark work although it has some limitations. While the pooling down-sampling operation expands the images' receptive field and promotes the integration of contextual information, it degrades the resolution and loses important location information.

Different types of methods have since been proposed to overcome this limitation. Several Encoder-Decoder-based~\cite{ronneberger2015u, xiao2018weighted, sun2019high} structures combined multi-level information to fine segmentation. The concept of Feature Pyramid Network (FPN) was utilized in Pyramid Scene Parsing Network (PSPNet)~\cite{zhao2017pyramid}, a multi-scale estimating network with contextual information integration. Due to the importance of the receptive field for scene understanding, Dilated Convolution~\cite{yu2015multi} was applied to the DeepLab Family~\cite{chen2017deeplab, chen2017rethinking} for images semantic segmentation to produce a considerable performance improvement. Besides, attention-based~\cite{dosovitskiy2020image, lin2021attention, liu2021swin, cao2021swin} models capable of extracting long-range contextual information were introduced into the image segmentation task. However, the lack of 3D spatial geometry information in 2D images hinders semantic comprehension of scenes. The performance of scene understanding can be improved with information from other data domains, such as 3D point cloud representation.

\subsection{3D Semantic Segmentation}
To cope with the structuring problem of point cloud data, one popular research is to apply projection-based  techniques~\cite{lawin2017deep, boulch2017unstructured, tatarchenko2018tangent}. These efforts can seem as transforming 3D point clouds into regular 2D pixels that can be learned using mature 2D segmentation methods. Nevertheless, inherent drawbacks still exist. For example, the fact that the 2D image is only an approximation of the 3D scene leads to a loss of geometric structure. Meanwhile, multi-viewpoint projections are quite sensitive to the viewpoints chosen.

In order to deal with the unordered and unstructured problems arising in point clouds,  one popular method is to transform point clouds into discrete structured representations. Voxelised point clouds can be processed by 3D convolution~\cite{huang2016point, tchapmi2017segcloud} in the same way as pixels in 2D neural networks. The voxel-based approach preserves the geometric structure information of the 3D point cloud neighbourhood. However, the data volume of the point cloud is enormous, and high-resolution voxels result in high memory and computational costs, while lower resolutions cause loss of detail, thus reducing segmentation accuracy. Consequently, 3D sparse convolutional networks~\cite{graham20183d, choy20194d} is designed to overcome these computational inefficiencies.

In summary, while both projection-based and voxel-based approaches have some drawbacks, direct processing of point clouds to achieve semantic segmentation has become a popular research topic in recent years. PointNet~\cite{qi2017pointnet}, with shared Multi-Layer Perceptrons (MLPs) as a component, can extract point-wise features and capture the global features of the point cloud using max-pooling. Since PointNet was originally proposed without considering the local information between neighbouring point clouds, researchers proposed different solutions to capture the local structure of point clouds, such as PointNet++~\cite{qi2017pointnet++} and DGCNN~\cite{wang2019dynamic}. In addition, point-based methods include attention-based features enhancement~\cite{zhao2019pooling, zhao2021point}, graph-based relation extraction~\cite{landrieu2018large, wang2019graph}, point-convolution operators~\cite{wang2018deep, wu2019pointconv, thomas2019kpconv}. Furthermore, sparse 3D point clouds lack continuous texture detail information, resulting in limited recognition performance of 3D scenes, which can be improved by exploiting 2D information.

\subsection{3D Semantic Segmentation Based on Joint 2D-3D data}
There has been some research in recent years on 2D and 3D data fusion, which can be broken down into unidirectional projection networks~\cite{dai20183dmv, chiang2019unified, jaritz2019multi, kundu2020virtual} and bidirectional projection networks~\cite{hu2021bidirectional}. The bidirectional projection network, typified by BPNet~\cite{hu2021bidirectional}, focuses on both 2D and 3D semantic segmentation tasks. Due to the mutual flow of its 2D and 3D information, its network framework has to rely on a symmetrical decoding network. In addition to the inflexibility of its framework, the 2D task introduced by the bi-projection idea will distract the network from the 3D segmentation task. This is our motivation for choosing a framework based on the uni-projection idea.

The successful fusion of 2D texture information and 3D structure information depends on the choice of 2D view, the location of feature fusion, and the feature space alignment. In terms of view selection, 3DMV~\cite{dai20183dmv} and MVPNet~\cite{jaritz2019multi} adopted a scheme with a fixed number of views, which means the views may not cover the entire 3D scene, and thus, cutting the scene would harm the performance of 3D semantic segmentation. In contrast, VMFusion~\cite{kundu2020virtual} tackles narrow viewing angle and occlusion problems in a multi-view fusion method by establishing multiple virtual viewpoints to obtain multiple views in the 3D scene. This scheme, however, has a high computational overhead that increases with the number of views. With an excellent balance, our work employs a dynamic view scheme that selects views based on the greedy algorithm of MVPNet until the view covers more than 90\% of the 3D scene while keeping the scene uncut.

As for fusing 2D and 3D features, there are typically two schemes. MVPNet~\cite{jaritz2019multi} utilizes 2D CNN to extract deep semantic features to enhance the original 3D shallow features (XYZ \& RGB) and finally achieves 3D semantic segmentation through PointNet++~\cite{qi2017pointnet++}. This method is called early fusion because the features are fused before the 3D network. The difference is that BPNet~\cite{hu2021bidirectional} applies the Bidirectional Projection Network (BPM) module on the decoding side of the U-shaped structure to align the 2D and 3D features of each layer, and then the 2D and 3D features are fused and passed to the next layer through one convolutional layer. The fusion of this scheme happens in the middle of the network. Therefore it can be named intermediate fusion. Since we chose the unidirectional projection scheme, our model will be based on early fusion as in MVPNet~\cite{jaritz2019multi}. However, we argue that the direct fusion of deep semantics and shallow semantics will cause a misalignment of feature space. For this reason, we propose a new fusion framework, DMF-Net, which leverages 2D and 3D feature extractors to extract the corresponding deep semantic features. Subsequently, the aligned 2D and 3D features are concatenated and passed through a complete sparse 3D convolutional network to integrate the information of the two data domains to achieve 3D semantic segmentation.

\section{Methodology}
\subsection{Overview}
\begin{figure}
	\centering
	\includegraphics[width=1\textwidth]{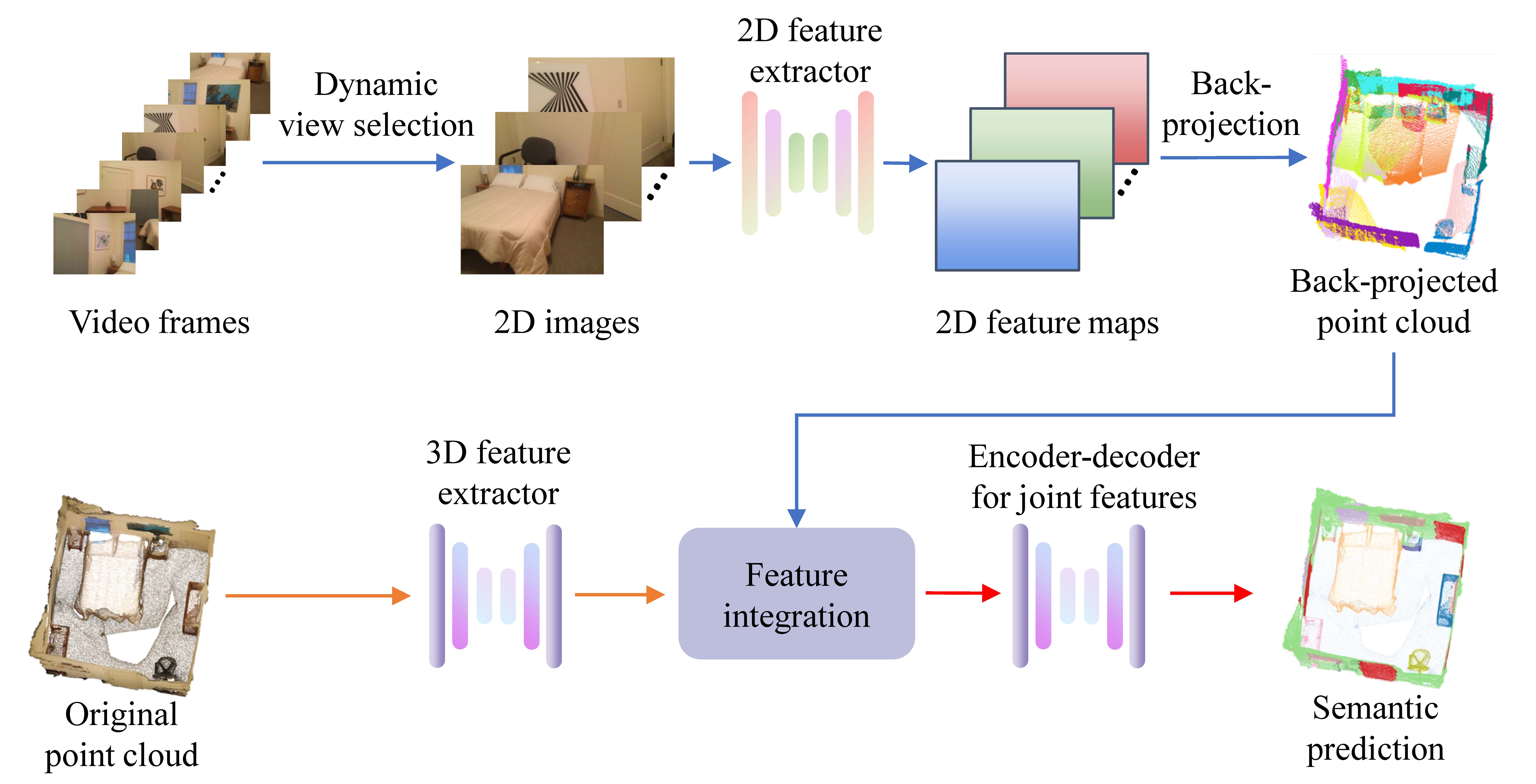}
	\caption{Overview of the proposed DMF-Net}
	\label{fig:Overview of the DMF-Net}
\end{figure}

An overview of our DMF-Net pipeline  is illustrated in Figure~\ref{fig:Overview of the DMF-Net}. Each scene data consist of a sequence of video frames and one point cloud scene. The input point cloud is called the original point cloud, while the point cloud formed by back-projecting all 2D feature maps into 3D space is called a back-projected point cloud with 2D features.
DMF-Net consists of three U-shaped sub-networks, where the 2D feature extractor is 2D U-Net~\cite{ronneberger2015u}, and the 3D feature extractor is 3D MinkowskiUNet~\cite{choy20194d}. Moreover, the encoder-decoder for joint features is also the 3D MinkowskiUNet. Although we set a specific network backbone in our implementation, different network backbones can also be utilised in our DMF-Net. The view selection and back-projection modules will be elaborated in Sec.~\ref{subsec:Dynamic View Selection} and Sec.~\ref{subsec:Unidirectional Projection Module}, respectively. As for the feature integration module, similar to MVPNet~\cite{jaritz2019multi}, it finds $k$ nearest neighbour 2D features for each point of the original point cloud. Subsequently, it directly concatenates with the deep 3D semantic features of the original point cloud and input to 3D MinkowskiUNet for further learning to predict the semantic results of the entire scene.

\subsection{Dynamic View Selection}
\label{subsec:Dynamic View Selection}
Previous work fixed the number of views, which would however cause insufficient overlaps between all the back-projected RGB-D frames and the scene point cloud. To make all the back-projected images cover as much of the scene point cloud as possible, many methods, e.g. MVPNet~\cite{jaritz2019multi}, generally choose to cut the point cloud scene. This will affect the recognition accuracy of cutting-edge objects. To this end, we propose a method for dynamic view selection, which sets a threshold for overlaps to ensure that the scene coverage is greater than 90\% so that the number of views selected for each scene is different. As for the calculation of overlaps, we follow the method of MVPNet. Finally, the greedy algorithm dynamically selects the appropriate number of views for each scene.

\subsection{Unidirectional Projection Module}
\label{subsec:Unidirectional Projection Module}
The video frames in the benchmark ScanNetv2~\cite{dai2017scannet} dataset used in our experiments are captured by a fixed camera and reconstructed into a 3D scene point cloud. Therefore, we establish a mapping relationship between multi-view images and 3D point clouds based on depth maps, camera intrinsics, and poses.  The world coordinate system is located where the point cloud scene is located, while the multi-view pictures belong to the pixel coordinate system. $\left(x_{w}, y_{w}, z_{w}\right)^{T}$ denotes a point in the world coordinate system and $(u, v) ^{T}$ denotes a pixel point in the pixel coordinate system. Thus, the formula for converting the pixel coordinate system to the world coordinate system is shown as follows.
\begin{equation}
	\left[\begin{array}{c}
		x_{w} \\
		y_{w} \\
		z_{w} \\
		1
	\end{array}\right]=Z_{c}K^{-1}\left[\begin{array}{cc}
		R & t \\
		0^{T} & 1
	\end{array}\right]^{-1}\left[\begin{array}{c}
		u \\
		v \\
		1 
	\end{array}\right] \;,
	\label{eq:3.1}
\end{equation}
where $Z_{c}$ is the depth value of the image, $K$ is the camera internal parameter matrix, $R$ is the orthogonal rotation matrix, and $t$ is the translation vector. 

\begin{figure} [ht]
	\centering
	\includegraphics[width=1.02\textwidth]{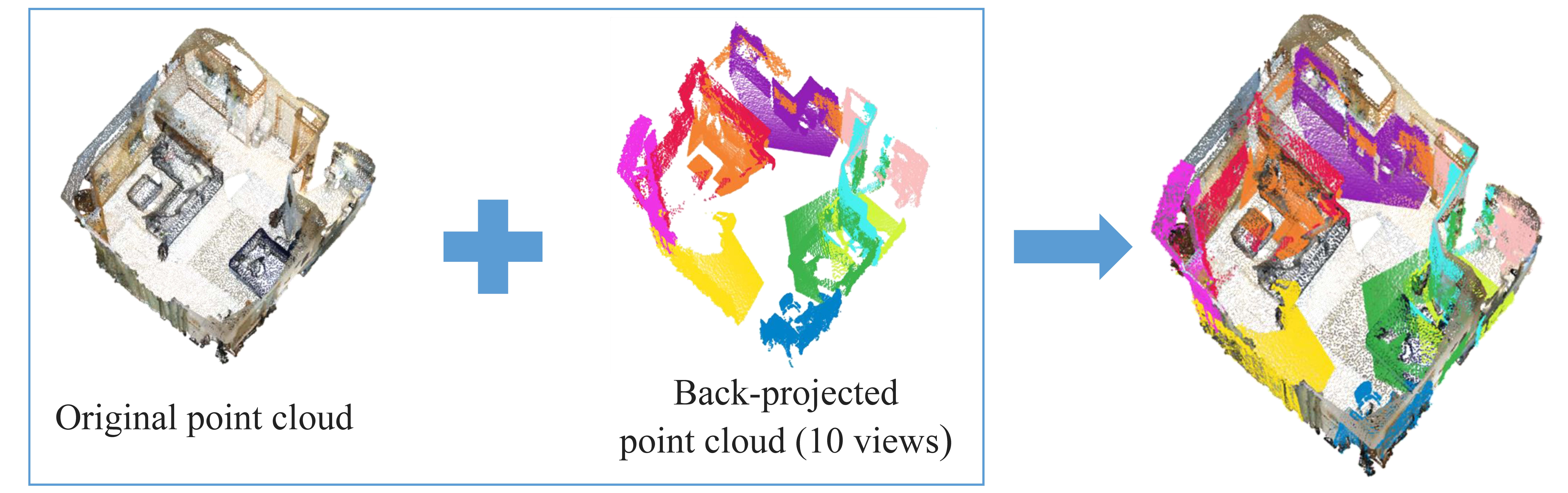}
	\caption{Back-projection visualisation result}
	\label{fig:Back-projection Visualisation Result}
\end{figure}

To verify our unidirectional projection module, we back-project all the dynamically selected multi-view images into 3D space and put them together with the original point cloud. The visualization results are shown in Figure~\ref{fig:Back-projection Visualisation Result}. Each color in the back-projected point cloud represents the projected point set for each view. It is clear to see that all views dynamically selected basically cover the indoor objects in the whole point cloud scene.

\subsection{Feature Integration}
\label{subsec:Feature Integration}
\begin{figure}[H]
	\centering
	\includegraphics[width=1\textwidth]{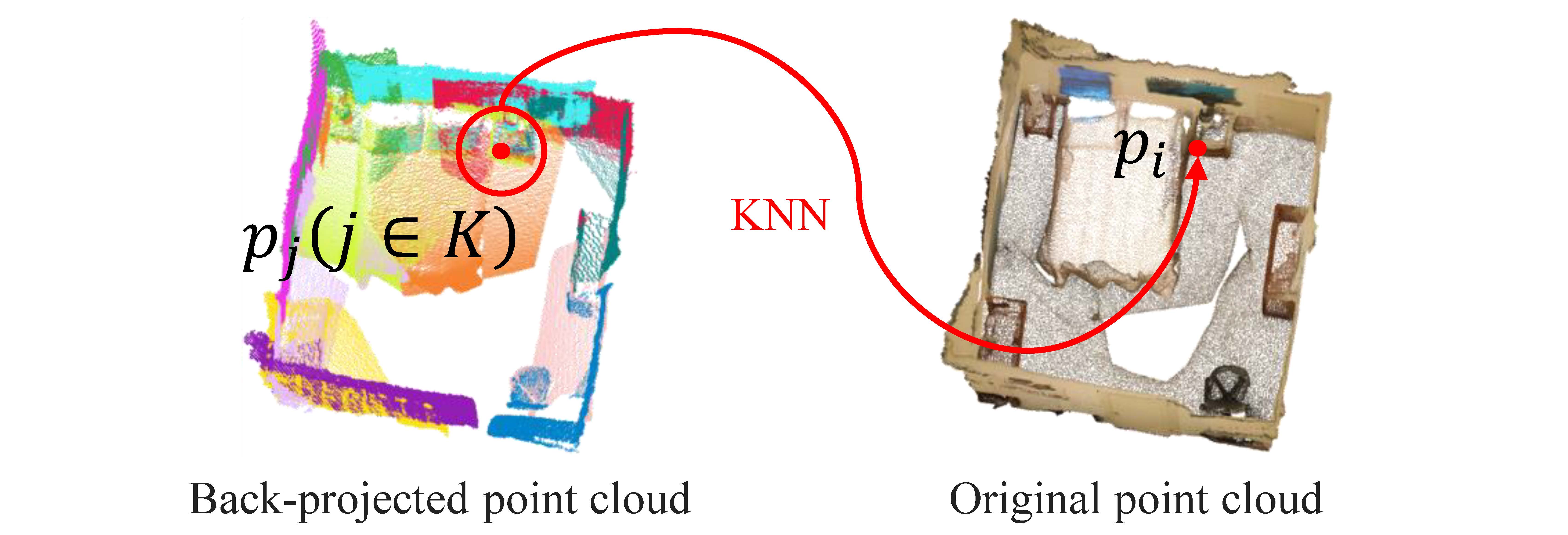}
	\caption{Feature integration method}
	\label{fig:Feature Integration Method}
\end{figure}

The multi-view images are mapped to the space of the original point cloud using the unidirectional projection module in Sec.~\ref{subsec:Unidirectional Projection Module} to obtain the back-projected point cloud, each point of which contains 64-dimensional 2D deep semantic feature information, denoted as $f_{j}$. Here, we define each point of the original point cloud as $p_{i}(x,y,z)$, as shown in Figure~\ref{fig:Feature Integration Method}. Specifically, each $p_{i}$ utilizes the $k$ Nearest Neighbor (KNN) algorithm to find $k$ back-projected points $p_{j}(j \in K)$ in the 3D Euclidean distance space. $f_{j}$ is summed to obtain $f_{2d}$ representing the 2D features of the $p_{i}$, while $p_{i}$ has obtained the 64-dimensional 3D deep semantic feature $f_{3d}$ through the 3D feature extractor. Finally, $f_{2d}$ and $f_{3d}$ are directly concatenated to become a 128-dimensional fusion feature $F_{i}$, which  is calculated as follows
\begin{equation}
    F_{i}=\text{Concat}\left[ f_{2d}, f_{3d}\right]
	\label{eq:3.2}
\end{equation}
\begin{equation}
    f_{2d}=\sum_{j \in N_{k}(i)} f_{j}
	\label{eq:3.3}
\end{equation}

\section{Experiments}
\label{sec:exp}
\subsection{Dataset and Metric}
The proposed model has been evaluated on ScanNetv2~\cite{dai2017scannet} dataset, which includes indoor scenes such as bedrooms, offices, and living rooms. The dataset captures 2.5 million RGB-D frames in 706 different scenes, officially divided into 1,201 training and 312 validation scans. Each scene is scanned around one to three times. Besides, an extra test set of 100 scans with hidden ground truth is used for the benchmark. The official evaluation metric of the Benchmark is the mean of class-wise intersection over union (mIoU).

\subsection{Implementation Details}
The training process can be divided into two stages. In the first stage, 2D images of ScanNetv2 were utilized to train a 2D feature extractor with 2D semantic labels. It is noted that the original image resolution was downsampled to $320\times 240$ for model acceleration and memory savings. In the second stage, a 3D network was trained with the frozen 2D feature extractor. Cross-Entropy loss was employed to predict 3D semantic labels. As for the hyperparameter $k$ in the feature integration module, we followed the previous practice, e.g. MVPNet and set it to 3. In the ablation study, the network structure of the proposed 3D feature extractor was set to MinkowskiUNet18A and the voxel size was set to 5 cm, which is consistent with the BPNet setup for a fair comparison. DMF-Net was trained for 500 epochs using the Adam~\cite{kingma2014adam} optimizer. The initial learning rate was set to 0.001, which decays with the cosine anneal schedule~\cite{loshchilov2016sgdr} at the 150th epochs. Besides, we conduct training with two Quadro RTX8000 cards and a mini-batch size of 16. 

\subsection{Comparison with SoTAs on ScanNetv2 Benchmark}
\subsubsection{Quantitative results}
We compare our method with mainstream methods on the test set of ScanNetv2 to evaluate the 3D semantic segmentation performance of DMF-Net. The majority of these methods can be divided into point-based methods (\cite{qi2017pointnet++}\cite{thomas2019kpconv}), convolution-based methods (\cite{zhang2020fusion}\cite{lin2020fpconv}\cite{wu2019pointconv}\cite{choy20194d}), and 2D-3D fusion-based methods (\cite{su2018splatnet}\cite{dai20183dmv}\cite{jaritz2019multi}\cite{hu2021bidirectional}). The results are reported in Table~\ref{table:4.1}.

\begin{table}[htbp]
	\begin{center}
    	\begin{adjustbox}{width=1\textwidth}
    		\begin{tabular}{l|c|cccccccccccccccccccl}
    			\toprule
    			Method & mIOU & \rotatebox{90}{bathtub} & \rotatebox{90}{bed} & \rotatebox{90}{bookshelf} & \rotatebox{90}{cabinet} & \rotatebox{90}{chair} & \rotatebox{90}{counter} & \rotatebox{90}{curtain} & \rotatebox{90}{desk} & \rotatebox{90}{door} & \rotatebox{90}{floor} & \rotatebox{90}{otherfurniture} & \rotatebox{90}{picture} &  \rotatebox{90}{refrigerator} & \rotatebox{90}{shower curtain} & \rotatebox{90}{sink} & \rotatebox{90}{sofa} & \rotatebox{90}{table} & \rotatebox{90}{toilet} & \rotatebox{90}{wall} & \rotatebox{90}{window}\\
    			\hline
    			PointNet++~\cite{qi2017pointnet++} & 33.9 & 58.4 & 47.8 & 45.8 & 25.6 & 36.0 & 25.0 & 24.7 & 27.8 & 26.1 & 67.7 & 18.3 & 11.7 & 21.2 & 14.5 & 36.4 & 34.6 & 23.2 &   54.8 & 52.3 & 25.2\\
    			SPLATNet*~\cite{su2018splatnet} & 39.3 & 47.2 & 51.1 & 60.6 & 31.1 & 65.6 & 24.5 & 40.5 & 32.8 & 19.7 & 92.7 & 22.7 & 00.0 & 00.1 & 24.9 & 27.1 & 51.0 & 38.3 & 59.3 & 69.9 & 26.7 \\
    			3DMV*~\cite{dai20183dmv} & 48.4 & 48.4 & 53.8 & 64.3 & 42.4 & 60.6 & 31.0 & 57.4 & 43.3 & 37.8 & 79.6 & 30.1 & 21.4 & 53.7 & 20.8 & 47.2 & 50.7 & 41.3 & 69.3 & 60.2 & 53.9\\
    			FAConv~\cite{zhang2020fusion} & 63.0 & 60.4 & 74.1 & 76.6 & 59.0 & 74.7 & 50.1 & 73.4 & 50.3 & 52.7 & 91.9 & 45.4 & 32.3 & 55.0 & 42.0 & 67.8 & 68.8 & 54.4 & 89.6 & 79.5 & 62.7\\
    			FPConv~\cite{lin2020fpconv} & 63.9 & 78.5 & 76.0 & 71.3 & 60.3 & 79.8 & 39.2 & 53.4 & 60.3 & 52.4 & 94.8 & 45.7 & 25.0 & 53.8 & 72.3 & 59.8 & 69.6 & 61.4 & 87.2 & 79.9 & 56.7\\
    			MVPNet*~\cite{jaritz2019multi} & 64.1 & 83.1 & 71.5 & 67.1 & 59.0 & 78.1 & 39.4 & 67.9 & 64.2 & 55.3 & 93.7 & 46.2 & 25.6 & 64.9 & 40.6 & 62.6 & 69.1 & 66.6 & 87.7 & 79.2 & 60.8\\
    			PointConv~\cite{wu2019pointconv} & 66.6 & 78.1 & 75.9 & 69.9 & 64.4 & 82.2 & 47.5 & 77.9 & 56.4 & 50.4 & 95.3 & 42.8 & 20.3 & 58.6 & 75.4 & 66.1 & 75.3 & 58.8 &  90.2 & 81.3 & 64.2\\
    			KP-FCNN~\cite{thomas2019kpconv} & 68.4 & 84.7 & 75.8 & 78.4 & 64.7 & 81.4 & 47.3 & 77.2 & 60.5 & 59.4 & 93.5 & 45.0 & 18.1 & 58.7 & 80.5 & 69.0 & 78.5 & 61.4 & 88.2 & 81.9 & 63.2\\
    			MinkowskiNet~\cite{choy20194d} & 73.6 & 85.9 & 81.8 & \textbf{83.2} & 70.9 & \textbf{84.0} & 52.1 & 85.3 & 66.0 & 64.3 & 95.1 & 54.4 & 28.6 & 73.1 & \textbf{89.3} & 67.5 & 77.2 & 68.3 & 87.4 & 85.2 & 72.7\\
    			BPNet*~\cite{hu2021bidirectional} & 74.9 & \textbf{90.9 }& \textbf{81.8} & 81.1 & \textbf{75.2} & 83.9 & 48.5 & 84.2 & 67.3 & \textbf{64.4} & 95.7 & 52.8 & 30.5 & 77.3 & 85.9 & \textbf{78.8} & \textbf{81.8} & 69.3 & 91.6 & \textbf{85.6} & 72.3\\
    			\hline
    			DMF-Net(Ours)* & \textbf{75.2} & 90.6 & 79.3 & 80.2 & 68.9 & 82.5 & \textbf{55.6} & \textbf{86.7} & \textbf{68.1} & 60.2 & \textbf{96.0} & \textbf{55.5} & \textbf{36.5} & \textbf{77.9} & 85.9 & 74.7 & 79.5 & \textbf{71.7} & \textbf{91.7} & \textbf{85.6} & \textbf{76.4}\\
    			\bottomrule
    	    \end{tabular}
		\end{adjustbox}
	\end{center}
	\caption{Comparison with typical approaches on ScanNetv2 benchmark, including point-based, convolution-based and 2D-3D fusion-based (marked with *) methods}
	\label{table:4.1}
\end{table}

DMF-Net achieves a significant mIOU performance improvement compared with point-based methods which are limited by their receptive field range and inefficient local information extraction. For convolution-based methods, such as stronger sparse convolution, MinkowskiNet can expand the range of receptive fields. Our method outperforms MinkowskiNet by a relative 2.2\% on mIOU because 2D texture information was utilized. DMF-Net shows a relative improvement of 17.3\% on mIOU compared to MVPNet, a baseline unidirectional projection scheme. Such improvement can be attributed to the fact that the feature alignment problem of MVPNet is alleviated. More importantly, our unidirectional projection scheme DMF-Net is significantly better than the bidirectional projection method BPNet, one state-of-the-art in 2D-3D information fusion. The inflexibility of the BPNet framework limits its performance, while the high flexibility of our network framework enables further improvements. Moreover, further examination of the bidirectional projection idea based on DMF-Net indicates that over-fitting usually occurs, which will be detailed in the subsequent ablation experiments.

\subsubsection{Qualitative Results}
\begin{figure}[h]
	\centering
	\includegraphics[width=1\textwidth]{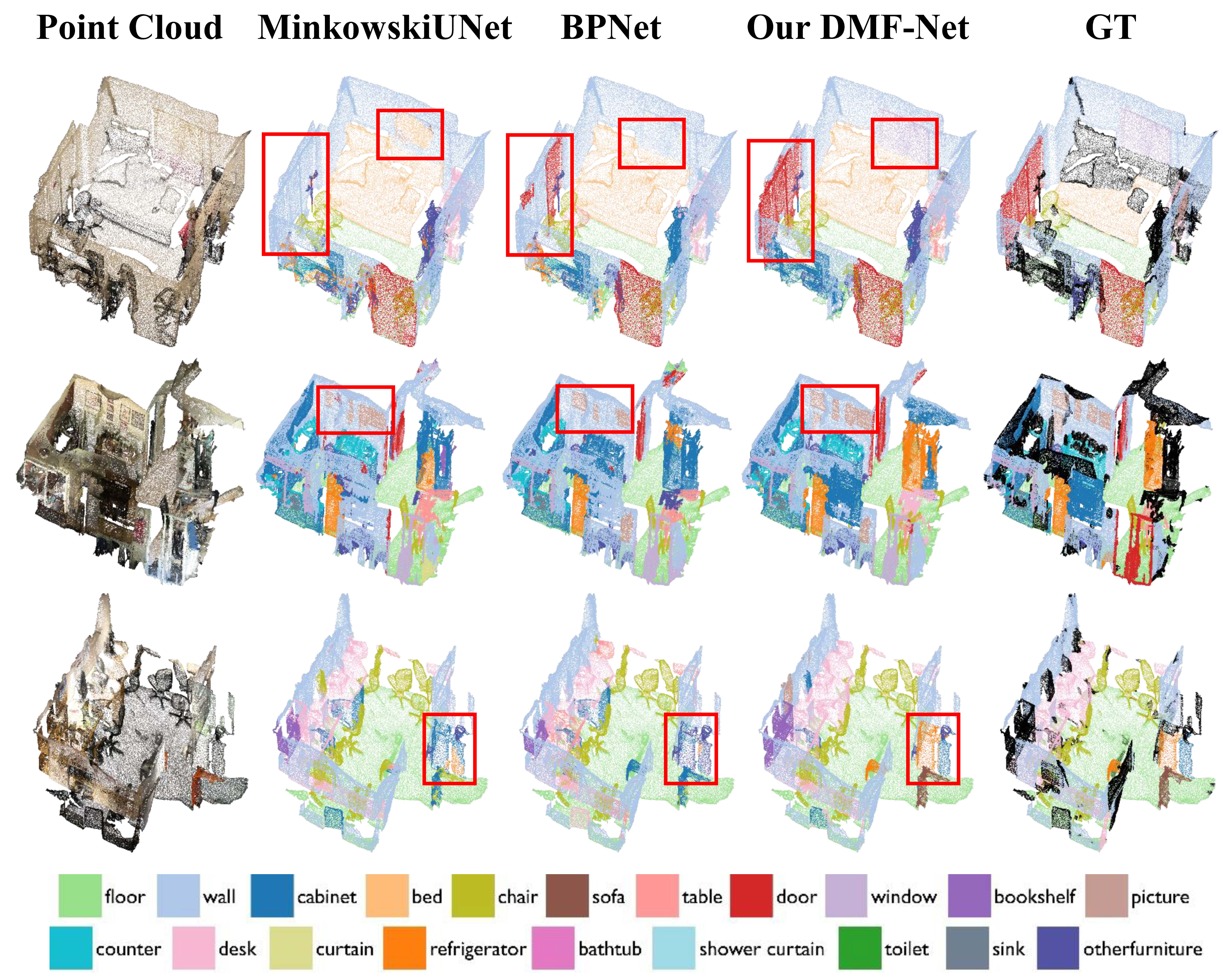}
	\caption{Qualitative results of 3D semantic segmentation}
	\label{fig:Qualitative Results of 3D Semantic Segmentation}
\end{figure}

We compare the pure 3D sparse convolution MinkowskiUNet, the joint 2D-3D approach BPNet, and our method DMF-Net to conduct inference on the validation set of ScanNetv2. The visualization results are shown in Figure~\ref{fig:Qualitative Results of 3D Semantic Segmentation}. 

As indicated by the red boxes, the 3D-only method MinkowskiUNet does not discriminate well between smooth planes or objects with insignificant shape differences, such as windows, doors, pictures, and refrigerators. This may due to the low resolution of the 3D point cloud and the lack of texture information for smooth planes. Despite the joint 2D-3D approach used by BPNet, the segmented objects are usually incomplete, owing that the bidirectional projection network distracts the core task, i.e. the 3D semantic segmentation.

\subsection{Ablation Study and Analysis}
\subsubsection{Ablation for 2D-3D Fusion Effectiveness}
We first fuse 2D deep semantic information with 3D shallow semantic information (i.e. each point contains position XYZ and clour RGB), followed by 3D sparse convolution. As shown in Table~\ref{table:4.2}, the 3D semantic segmentation performance mIOU is improved from 66.4 to 70.8, indicating that 2D semantic information can benefit the 3D semantic segmentation task. The direct fusion of 2D deep semantic features with 3D shallow geometric features will cause misalignment in the semantic depth space affecting the network's performance. For this reason, our DMF-Net adds a 3D feature extractor based on the above framework so that 2D \& 3D features are fused and aligned in semantic depth. As shown in Table~\ref{table:4.2}, the feature-aligned model (V2) has a relative improvement of 1.3\% on mIOU performance compared to the unaligned model (V1). In addition, we get a relative 4.4\% on mIOU improvement with the stronger attention-based 2D backbone Swin-UNet~\cite{cao2021swin} (V3) compared with the common U-Net34 model (V2). It is worth mentioning that the voxel size is sensitive to the performance of 3D sparse convolution. We adopt a deeper 3D sparse network, MinkowskiUNet34C, and set the voxel size to 2cm to obtain better results, as V4 shown in Table~\ref{table:4.2}.

\begin{table}[htbp]
    \begin{center}
        \begin{adjustbox}{width=0.65\textwidth}
            \begin{tabular}{l|c|cc}
            \toprule
            \multirow{2}{*}{Method}            & \multicolumn{1}{c|}{\multirow{2}{*}{Voxel Size}} & \multicolumn{2}{c}{mIOU}                             \\ \cline{3-4} 
                                               & \multicolumn{1}{c|}{}                            & \multicolumn{1}{c}{2D}   & 3D                        \\ \hline
            U-Net34~\cite{ronneberger2015u}                            & -                                                & \multicolumn{1}{c}{60.7} & -                         \\ 
            Swin-UNet~\cite{cao2021swin}                          & -                                                & \multicolumn{1}{c}{68.8} & -                         \\ 
            MinkowskiUNet18A~\cite{choy20194d}                   & 5cm                                              & \multicolumn{1}{c}{-}    & 66.4                      \\ \hline
            Ours V1(U-Net34 + XYZ \& RGB)  & 5cm                                              & \multicolumn{1}{c}{-}    & \multicolumn{1}{c}{69.9} \\ 
            Ours V2(U-Net34 + MinkowskiUNet18A)  & 5cm                                              & \multicolumn{1}{c}{-}    & \multicolumn{1}{c}{70.8} \\ 
            Ours V3(Swin-UNet + MinkowskiUNet18A) & 5cm                                              & \multicolumn{1}{c}{-}    & \multicolumn{1}{c}{73.9} \\ 
            Ours V4(Swin-UNet + MinkowskiUNet34C) & 2cm                                              & \multicolumn{1}{c}{-}    & \multicolumn{1}{c}{\textbf{75.6}} \\ 
            \bottomrule
            \end{tabular}
        \end{adjustbox}
    \end{center}
	\caption{2D$\And$3D semantic segmentation on the validation set of ScanNetv2}
	\label{table:4.2}
\end{table}

\subsubsection{Ablation for Projection Methods}
We conduct further ablative experiments to verify that the unidirectional projection scheme is more focused on the 3D semantic segmentation task than the bidirectional projection. Using the same framework in Figure~\ref{fig:Overview of the DMF-Net}, we also project 3D features into the 2D deep semantic feature space. Essentially, we apply  a projection method similar to Sec.~\ref{subsec:Unidirectional Projection Module}, which is an opposite process. Meanwhile, the 2D-3D feature fusion is the same as in Sec.~\ref{subsec:Feature Integration}. After 3D features are fused with  multi-view features, semantic labels are output through a U-Net34 Network. Hence 2D cross-entropy loss is introduced on the total loss of the model. To avoid focusing too much on the optimization of 2D tasks, we multiply the 2D loss by a weight of 0.1, which is also the same as BPNet. At this time, the new 2D model parameters for training no longer freeze, and the learning rate of all 2D models is 10 times lower than that of the 3D model.

Our experiments show that the bidirectional projection model overfits  when the model is trained to the 200th epoch, as seen from the 3D validation loss in Figure~\ref{fig:Validation Loss for Unidirectional & Bidirectional Projection}. Meanwhile, the 3D mIOU of bidirectional projection only reaches 70.6, which is lower than the performance of the simple unidirectional projection (70.8). As  the 2D task is also introduced, the increased learning parameters and the difficulty in adjusting the hyperparameters made it difficult for the model to focus more on 3D tasks. In this sense, unidirectional projection can focus more on 3D semantic segmentation tasks than bidirectional projection, leading to better flexibility.

\subsection{DMF-Net on NYUv2}
To verify the generalization ability, we conduct experiments on another popular RGB-D dataset, NYUv2~\cite{silberman2012indoor}. The dataset contains 1,449 densely labelled pairs of aligned RGB and depth images. We follow the official split of the dataset, using 795 for training and 654 for testing. Since this dataset has no 3D data, we need to use depth and camera intrinsics to generate 3D point clouds with 2D labels. We report a dense pixel classification mean accuracy for DMF-Net, obtaining a significant performance improvement compared to other typical methods, especially joint 2D-3D methods, e.g. 3DMV~\cite{dai20183dmv} and BPNet~\cite{hu2021bidirectional}.
\begin{table}[htbp]
	\begin{center}
    	\begin{adjustbox}{width=0.4\textwidth}
        	\begin{tabular}{c|c}
        		\toprule
        		Method  & mean Accuracy\\
        		\hline
        		SceneNet~\cite{handa1511scenenet}  & 52.5\\
        		Hermans et al.~\cite{hermans2014dense}  & 54.3\\
        		SemanticFusion~\cite{mccormac2017semanticfusion}  & 59.2\\
        		Scannet~\cite{dai2017scannet} & 60.7\\
        		3DMV~\cite{dai20183dmv} & 71.2\\
                BPNet~\cite{hu2021bidirectional} & 73.5\\
                \hline
                \textbf{DMF-Net} & \textbf{78.4}\\
        		\bottomrule
        	\end{tabular}
        \end{adjustbox}
    \end{center}
	\caption{Semantic segmentation results (13-class task) on NYUv2~\cite{silberman2012indoor}}
	\label{table:4.3}
\end{table}

\section{Conclusion}
\label{sec:con}
In our work, we propose a Deep Multi-view Fusion Network (DMF-Net) based on a unidirectional projection method to perform 3D semantic segmentation utilizing 2D continuous texture information and 3D geometry information. Compared with the previous 2D-3D fusion methods, DMF-Net enjoys a deeper and more flexible network. Thus DMF-Net enables improved segmentation accuracy for objects with little variation in shape, effectively compensating for the limitations of pure 3D methods. In addition, DMF-Net achieves the superior performance of the joint 2D-3D method in the ScanNetv2 benchmark. Moreover, we obtain significant performance gains over previous approaches on the NYUv2 dataset.

Currently,  the number of dynamically selected multi-view images in DMF-Net is relatively large in order to cover the full 3D scene.  In the future, we will explore efficient view selection algorithms so that even a few image inputs could  achieve the full  coverage of the 3D scene. 

\section*{Acknowledgements}
The work was partially supported by the following: National Natural Science Foundation of China under no.61876155; Jiangsu Science and Technology Programme (Natural Science Foundation of Jiangsu Province) under no.  BE2020006-4, UK Engineering and Physical Sciences Research Council (EPSRC) Grants Ref. EP/M026981/1, EP/T021063/1, EP/T024917/.



\bibliographystyle{elsarticle-num} 
\bibliography{references}

\begin{thebibliography}{10}
\expandafter\ifx\csname url\endcsname\relax
  \def\url#1{\texttt{#1}}\fi
\expandafter\ifx\csname urlprefix\endcsname\relax\def\urlprefix{URL }\fi
\expandafter\ifx\csname href\endcsname\relax
  \def\href#1#2{#2} \def\path#1{#1}\fi

\bibitem{FERNANDES2021161}
D.~Fernandes, A.~Silva, R.~Névoa, C.~Simões, D.~Gonzalez, M.~Guevara,
  P.~Novais, J.~Monteiro, P.~Melo-Pinto, Point-cloud based 3d object detection
  and classification methods for self-driving applications: A survey and
  taxonomy, Information Fusion 68 (2021) 161--191.

\bibitem{szegedy2015going}
C.~Szegedy, W.~Liu, Y.~Jia, P.~Sermanet, S.~Reed, D.~Anguelov, D.~Erhan,
  V.~Vanhoucke, A.~Rabinovich, Going deeper with convolutions, in: CVPR, 2015,
  pp. 1--9.

\bibitem{he2016deep}
K.~He, X.~Zhang, S.~Ren, J.~Sun, Deep residual learning for image recognition,
  in: CVPR, 2016, pp. 770--778.

\bibitem{yu2015multi}
F.~Yu, V.~Koltun, Multi-scale context aggregation by dilated convolutions, in:
  ICLR, 2016.

\bibitem{zhao2017pyramid}
H.~Zhao, J.~Shi, X.~Qi, X.~Wang, J.~Jia, Pyramid scene parsing network, in:
  CVPR, 2017, pp. 2881--2890.

\bibitem{yao2021scaffold}
K.~Yao, K.~Huang, J.~Sun, L.~Jing, D.~Huang, C.~Jude, Scaffold-a549: a
  benchmark 3d fluorescence image dataset for unsupervised nuclei segmentation,
  Cognitive Computation 13~(6) (2021) 1603--1608.

\bibitem{qi2017pointnet}
C.~R. Qi, H.~Su, K.~Mo, L.~J. Guibas, Pointnet: Deep learning on point sets for
  3d classification and segmentation, in: CVPR, 2017, pp. 652--660.

\bibitem{qi2017pointnet++}
C.~R. Qi, L.~Yi, H.~Su, L.~J. Guibas, Pointnet++ deep hierarchical feature
  learning on point sets in a metric space, in: NIPS, 2017, pp. 5105--5114.

\bibitem{wu2019pointconv}
W.~Wu, Z.~Qi, L.~Fuxin, Pointconv: Deep convolutional networks on 3d point
  clouds, in: CVPR, 2019, pp. 9621--9630.

\bibitem{jiang2023pointgs}
C.~Jiang, K.~Huang, J.~Wu, X.~Wang, J.~Xiao, A.~Hussain, Pointgs: Bridging and
  fusing geometric and semantic space for 3d point cloud analysis, Information
  Fusion 91 (2023) 316--326.

\bibitem{graham20183d}
B.~Graham, M.~Engelcke, L.~Van Der~Maaten, 3d semantic segmentation with
  submanifold sparse convolutional networks, in: CVPR, 2018, pp. 9224--9232.

\bibitem{choy20194d}
C.~Choy, J.~Gwak, S.~Savarese, 4d spatio-temporal convnets: Minkowski
  convolutional neural networks, in: CVPR, 2019, pp. 3075--3084.

\bibitem{zhao2022divide}
W.~Zhao, Y.~Yan, C.~Yang, J.~Ye, X.~Yang, K.~Huang, Divide and conquer: 3d
  point cloud instance segmentation with point-wise binarization, arXiv
  preprint arXiv:2207.11209 (2022).

\bibitem{dai20183dmv}
A.~Dai, M.~Nie{\ss}ner, 3dmv: Joint 3d-multi-view prediction for 3d semantic
  scene segmentation, in: ECCV, 2018, pp. 452--468.

\bibitem{chiang2019unified}
H.-Y. Chiang, Y.-L. Lin, Y.-C. Liu, W.~H. Hsu, A unified point-based framework
  for 3d segmentation, in: 3DV, 2019, pp. 155--163.

\bibitem{jaritz2019multi}
M.~Jaritz, J.~Gu, H.~Su, Multi-view pointnet for 3d scene understanding, in:
  ICCVW, 2019.

\bibitem{kundu2020virtual}
A.~Kundu, X.~Yin, A.~Fathi, D.~Ross, B.~Brewington, T.~Funkhouser,
  C.~Pantofaru, Virtual multi-view fusion for 3d semantic segmentation, in:
  ECCV, 2020, pp. 518--535.

\bibitem{hu2021bidirectional}
W.~Hu, H.~Zhao, L.~Jiang, J.~Jia, T.-T. Wong, Bidirectional projection network
  for cross dimension scene understanding, in: CVPR, 2021, pp. 14373--14382.

\bibitem{dai2017scannet}
A.~Dai, A.~X. Chang, M.~Savva, M.~Halber, T.~Funkhouser, M.~Nie{\ss}ner,
  Scannet: Richly-annotated 3d reconstructions of indoor scenes, in: CVPR,
  2017, pp. 5828--5839.

\bibitem{cao2021swin}
H.~Cao, Y.~Wang, J.~Chen, D.~Jiang, X.~Zhang, Q.~Tian, M.~Wang, Swin-unet:
  Unet-like pure transformer for medical image segmentation, arXiv preprint
  arXiv:2105.05537 (2021).

\bibitem{silberman2012indoor}
N.~Silberman, D.~Hoiem, P.~Kohli, R.~Fergus, Indoor segmentation and support
  inference from rgbd images, in: ECCV, 2012, pp. 746--760.

\bibitem{ronneberger2015u}
O.~Ronneberger, P.~Fischer, T.~Brox, U-net: Convolutional networks for
  biomedical image segmentation, in: MICCAI, 2015, pp. 234--241.

\bibitem{huang2019deep}
K.~Huang, A.~Hussain, Q.~Wang, R.~Zhang, Deep learning: fundamentals, theory
  and applications, Vol.~2, Springer, 2019.

\bibitem{long2015fully}
J.~Long, E.~Shelhamer, T.~Darrell, Fully convolutional networks for semantic
  segmentation, in: CVPR, 2015, pp. 3431--3440.

\bibitem{xiao2018weighted}
X.~Xiao, S.~Lian, Z.~Luo, S.~Li, Weighted res-unet for high-quality retina
  vessel segmentation, in: ITME, 2018, pp. 327--331.

\bibitem{sun2019high}
K.~Sun, B.~Xiao, D.~Liu, J.~Wang, Deep high-resolution representation learning
  for human pose estimation, in: CVPR, 2019, pp. 5693--5703.

\bibitem{chen2017deeplab}
L.-C. Chen, G.~Papandreou, I.~Kokkinos, K.~Murphy, A.~L. Yuille, Deeplab:
  Semantic image segmentation with deep convolutional nets, atrous convolution,
  and fully connected crfs, IEEE Transactions on Pattern Analysis and Machine
  Intelligence (TPAMI) 40~(4) (2017) 834--848.

\bibitem{chen2017rethinking}
L.-C. Chen, G.~Papandreou, F.~Schroff, H.~Adam, Rethinking atrous convolution
  for semantic image segmentation, arXiv preprint arXiv:1706.05587 (2017).

\bibitem{dosovitskiy2020image}
A.~Dosovitskiy, L.~Beyer, A.~Kolesnikov, D.~Weissenborn, X.~Zhai,
  T.~Unterthiner, M.~Dehghani, M.~Minderer, G.~Heigold, S.~Gelly, J.~Uszkoreit,
  N.~Houlsby, An image is worth 16x16 words: Transformers for image recognition
  at scale, in: ICLR, 2021.

\bibitem{lin2021attention}
X.~Lin, G.~Zhong, K.~Chen, Q.~Li, K.~Huang, Attention-augmented machine memory,
  Cognitive Computation 13~(3) (2021) 751--760.

\bibitem{liu2021swin}
Z.~Liu, Y.~Lin, Y.~Cao, H.~Hu, Y.~Wei, Z.~Zhang, S.~Lin, B.~Guo, Swin
  transformer: Hierarchical vision transformer using shifted windows, in: ICCV,
  2021, pp. 9992--10002.

\bibitem{lawin2017deep}
F.~J. Lawin, M.~Danelljan, P.~Tosteberg, G.~Bhat, F.~S. Khan, M.~Felsberg, Deep
  projective 3d semantic segmentation, in: CAIP, 2017, pp. 95--107.

\bibitem{boulch2017unstructured}
A.~Boulch, B.~Le~Saux, N.~Audebert, Unstructured point cloud semantic labeling
  using deep segmentation networks, in: 3DOR, 2017.

\bibitem{tatarchenko2018tangent}
M.~Tatarchenko, J.~Park, V.~Koltun, Q.-Y. Zhou, Tangent convolutions for dense
  prediction in 3d, in: CVPR, 2018, pp. 3887--3896.

\bibitem{huang2016point}
J.~Huang, S.~You, Point cloud labeling using 3d convolutional neural network,
  in: ICPR, 2016, pp. 2670--2675.

\bibitem{tchapmi2017segcloud}
L.~Tchapmi, C.~Choy, I.~Armeni, J.~Gwak, S.~Savarese, Segcloud: Semantic
  segmentation of 3d point clouds, in: 3DV, 2017, pp. 537--547.

\bibitem{wang2019dynamic}
Y.~Wang, Y.~Sun, Z.~Liu, S.~E. Sarma, M.~M. Bronstein, J.~M. Solomon, Dynamic
  graph cnn for learning on point clouds, ACM Transactions On Graphics (TOG)
  38~(5) (2019) 1--12.

\bibitem{zhao2019pooling}
C.~Zhao, W.~Zhou, L.~Lu, Q.~Zhao, Pooling scores of neighboring points for
  improved 3d point cloud segmentation, in: ICIP, 2019, pp. 1475--1479.

\bibitem{zhao2021point}
H.~Zhao, L.~Jiang, J.~Jia, P.~H. Torr, V.~Koltun, Point transformer, in: ICCV,
  2021, pp. 16259--16268.

\bibitem{landrieu2018large}
L.~Landrieu, M.~Simonovsky, Large-scale point cloud semantic segmentation with
  superpoint graphs, in: CVPR, 2018, pp. 4558--4567.

\bibitem{wang2019graph}
L.~Wang, Y.~Huang, Y.~Hou, S.~Zhang, J.~Shan, Graph attention convolution for
  point cloud semantic segmentation, in: CVPR, 2019, pp. 10296--10305.

\bibitem{wang2018deep}
S.~Wang, S.~Suo, W.-C. Ma, A.~Pokrovsky, R.~Urtasun, Deep parametric continuous
  convolutional neural networks, in: CVPR, 2018, pp. 2589--2597.

\bibitem{thomas2019kpconv}
H.~Thomas, C.~R. Qi, J.-E. Deschaud, B.~Marcotegui, F.~Goulette, L.~J. Guibas,
  Kpconv: Flexible and deformable convolution for point clouds, in: ICCV, 2019,
  pp. 6411--6420.

\bibitem{kingma2014adam}
D.~P. Kingma, J.~Ba, Adam: A method for stochastic optimization, in: ICLR,
  2015.

\bibitem{loshchilov2016sgdr}
I.~Loshchilov, F.~Hutter, Sgdr: Stochastic gradient descent with warm restarts,
  in: ICLR, 2017.

\bibitem{zhang2020fusion}
J.~Zhang, C.~Zhu, L.~Zheng, K.~Xu, Fusion-aware point convolution for online
  semantic 3d scene segmentation, in: CVPR, 2020, pp. 4534--4543.

\bibitem{lin2020fpconv}
Y.~Lin, Z.~Yan, H.~Huang, D.~Du, L.~Liu, S.~Cui, X.~Han, Fpconv: Learning local
  flattening for point convolution, in: CVPR, 2020, pp. 4293--4302.

\bibitem{su2018splatnet}
H.~Su, V.~Jampani, D.~Sun, S.~Maji, E.~Kalogerakis, M.-H. Yang, J.~Kautz,
  Splatnet: Sparse lattice networks for point cloud processing, in: CVPR, 2018,
  pp. 2530--2539.

\bibitem{handa1511scenenet}
A.~Handa, V.~Patraucean, V.~Badrinarayanan, S.~Stent, R.~Cipolla, Scenenet:
  Understanding real world indoor scenes with synthetic data, arXiv preprint
  arXiv:1511.07041 (2015).

\bibitem{hermans2014dense}
A.~Hermans, G.~Floros, B.~Leibe, Dense 3d semantic mapping of indoor scenes
  from rgb-d images, in: ICRA, 2014, pp. 2631--2638.

\bibitem{mccormac2017semanticfusion}
J.~McCormac, A.~Handa, A.~Davison, S.~Leutenegger, Semanticfusion: Dense 3d
  semantic mapping with convolutional neural networks, in: ICRA, 2017, pp.
  4628--4635.

\end{thebibliography}






\end{document}